\documentclass[12pt]{article}
\usepackage{amsfonts}
\usepackage{a4wide}
\usepackage{graphicx}
\usepackage{amsmath,amscd,amsthm,a4,amssymb}

\begin{document}

\begin{center}
{\Large \textbf{On the universal approximation property of radial basis
function neural networks}}

\

{\large \textbf{Aysu Ismayilova}}

\smallskip

{School of Mathematics, University of Edinburgh}

Peter Guthrie Tait Road, Edinburgh EH9 3FD, Scotland

{e-mail:} a.ismayilova-1@sms.ed.ac.uk

\bigskip

{\large \textbf{Muhammad Ismayilov}}

\smallskip

{French-Azerbaijani University}

183 Nizami str., Baku, Azerbaijan

{e-mail:} mv.ismayilov@gmail.com
\end{center}

\smallskip

\textbf{Abstract.} In this paper we consider a new class of RBF (Radial
Basis Function) neural networks, in which smoothing factors are replaced
with shifts. We prove under certain conditions on the activation function
that these networks are capable of approximating any continuous multivariate
function on any compact subset of the $d$-dimensional Euclidean space. For
RBF networks with finitely many fixed centroids we describe conditions
guaranteeing approximation with arbitrary precision.

\smallskip


\smallskip


\

\begin{center}
{\large \textbf{1. Introduction}}
\end{center}

RBF (Radial Basis Function) neural networks are being used for function
approximation, time series forecasting, classification, pattern recognition
and system control problems. Besides their strong approximation capability,
these networks benefit from other powerful characteristics such as the
ability to represent complex nonlinear mappings and provide a fast and
robust learning mechanism without significant computational cost. The
literature abounds with different aspects and various applications of RBF
neural networks.

The fundamental principles and advantages of RBF neural networks were first
displayed in the papers of Broomhead and Lowe \cite{Br}, Moody and Darken
\cite{Moo}, Lipmann \cite{Lip} and Bishop \cite{Bish}. A variant of RBF
network with an input layer, a hidden layer and an output layer is
constructed by the following scheme. Each unit in the hidden layer of this
RBF network has its own centroid and for an input vector $\mathbf{x}%
=(x_{1},...,x_{d})$ it computes the distance between $\mathbf{x}$ and its
centroid $\mathbf{c}\in \mathbb{R}^{d}$. Its output (the output of a given
hidden unit) is some nonlinear function of that distance. Hence each hidden
unit computes a radial function, that is, a function which is constant on
the spheres $\left\Vert \mathbf{x}-\mathbf{c}\right\Vert =\alpha ,\;\alpha
\in \mathbb{R}$. Each output unit gives a weighted summation of the outputs
of hidden units. For the clarity of exposition, we will consider in the
sequel only a one dimensional output space instead of outputs represented by
multiple units. The generalization of our results to the $n$-dimensional
output space is straightforward.

Assuming that there are $d$ input units and one output unit, the final
response function has the following form:

\begin{equation*}
G(\mathbf{x})=\sum_{i=1}^{m}w_{i}g\left( \frac{\left\Vert \mathbf{x}-\mathbf{%
c}_{i}\right\Vert }{\sigma _{i}}\right) .\eqno(1.1)
\end{equation*}%
Here $m\in \mathbb{N}$ is the number of units in the hidden layer, $%
(w_{1},...,w_{m})\in \mathbb{R}^{m}$ is the vector of weights, $\mathbf{x}%
\in \mathbb{R}^{d}$ is an input vector, $\mathbf{c}_{i}\in \mathbb{R}^{d}$
and $\sigma _{i}\in \mathbb{R}_{+}$ are the centroids and smoothing factors
(or widths) of the $i$-th node, $1\leq i\leq m$, respectively, $\left\Vert
\mathbf{x}-\mathbf{c}_{i}\right\Vert $ is the Euclidean distance between $%
\mathbf{x}$ and $\mathbf{c}_{i}$, and $g:[0,+\infty )\rightarrow \mathbb{R}$
is the so-called activation function.

Various activation functions in RBF neural networks can be implemented and
the smoothing factors may be the same or may vary across units.

The RBF neural networks have the universal approximation property.
Theoretically, such networks can approximate any continuous multivariate
function within any degree of accuracy, if the activation function is
suitably chosen. The most well-known result is due to Park and Sandberg \cite%
{Park}. In 1993, they showed, along with other results, that for a
continuous and integrable $g(\left\Vert \mathbf{x}\right\Vert )$ (considered
as a function of $d$ variables) the set of functions (1.1) is dense in $C(%
\mathbb{R}^{d})$ in the topology of uniform convergence on compact subsets
of $\mathbb{R}^{d}$. That is, for any continuous function $f:\mathbb{R}%
^{d}\rightarrow \mathbb{R}$, for any compact subset $K\subset \mathbb{R}^{d}$
and for any $\varepsilon >0$, there exists a function $G$ of form (1.1) such
that

\begin{equation*}
\left\Vert f-G\right\Vert _{K}\overset{def}{=}\max_{\mathbf{x}\in
K}\left\vert f(\mathbf{x})-G(\mathbf{x})\right\vert <\varepsilon .
\end{equation*}%
The requirement of the integrability of $g(\left\Vert \mathbf{x}\right\Vert
) $ is relaxed in Liao, Fang and Nuttle \cite{Lia}. They showed that for an
activation function, which is continuous almost everywhere, locally
essentially bounded and nonpolynomial, the RBF networks (1.1) can
approximate any continuous function with arbitrary accuracy. There are also
other results on the universality of RBF neural networks (see, e.g., \cite%
{Hua,Lia,Par,Wu}).

In this paper we bring into consideration a new class of RBF neural
networks. In this class the smoothing factors $\sigma _{i}$ are replaced
with shifts $\nu _{i}\in \mathbb{R}$. That is, this class consists of
functions $H:\mathbb{R}^{d}\rightarrow \mathbb{R}$ of the form

\begin{equation*}
H(\mathbf{x})=\sum_{i=1}^{m}w_{i}g\left( \left\Vert \mathbf{x}-\mathbf{c}%
_{i}\right\Vert -\nu _{i}\right) .\eqno(1.2)
\end{equation*}

We are interested in the universal approximation property of such RBF neural
networks. For which activation functions $g$ functions of form (1.2) is
dense in $C(\mathbb{R}^{d})$ in the topology of uniform convergence on
compact subsets of $\mathbb{R}^{d}$. We will give various conditions on the
activation $g$ which guarantee the density of the functions (1.2) in $C(X)$
for any compact set $X\subset $ $\mathbb{R}^{d}$.

\bigskip

\bigskip

\bigskip

\bigskip

\begin{center}
{\large \textbf{2. Universal approximation theorems}}
\end{center}

The following theorem is based on the results of Park and Sandberg, and
Schwartz.

\bigskip

\textbf{Theorem 2.1.} \textit{Assume $d\geq 1,$ $1\leq p<\infty .$ Assume an
activation function $g\in C(\mathbb{R})\cap L^{p}(\mathbb{R})$ and $%
t^{d-1}g(t)$ is integrable on $[0,+\infty ).$ Then for any continuous
function $f:\mathbb{R}^{d}\rightarrow \mathbb{R}$, for any $\varepsilon >0$
and for any compact subset $X\subset \mathbb{R}^{d}$, there exist $m\in
\mathbb{N}$, $w_{i},\nu _{i}\in \mathbb{R}$, $\mathbf{c}_{i}\in \mathbb{R}%
^{d}$ such that}

\begin{equation*}
\left\vert f(\mathbf{x})-\sum_{i=1}^{m}w_{i}g\left( \left\Vert \mathbf{x}-%
\mathbf{c}_{i}\right\Vert -\nu _{i}\right) \right\vert <\varepsilon \eqno%
(2.1)
\end{equation*}%
\textit{for all $\mathbf{x}\in X.$}

\bigskip

\textbf{Proof.} The result of Park and Sandberg (see \cite[Theorem 5]{Park})
says that if $K:\mathbb{R}^{d}\rightarrow \mathbb{R}$ is continuous,
integrable and radially symmetric with respect to the Euclidean norm, then
the functions of the form

\begin{equation*}
q(\mathbf{x})=\sum_{i=1}^{k}w_{i}K\left( \frac{\mathbf{x}-\mathbf{c}_{i}}{%
\sigma _{i}}\right) .\eqno(2.2)
\end{equation*}%
are dense in $C(X)$ for any compact set $X\subset \mathbb{R}^{d}$.

Consider the function $K(\mathbf{x})=g(\left\Vert \mathbf{x}\right\Vert ).$
Clearly, $K(\mathbf{x})$ is radially symmetric and since $t^{d-1}g(t)$ is
integrable on $[0,+\infty )$, $K(\mathbf{x})$ is integrable on $\mathbb{R}%
^{d}$. Thus this function satisfies the hypothesis of Park and Sandberg's
theorem. Assume any function $f\in C(\mathbb{R}^{d})$, any number $%
\varepsilon >0$ and any compact subset $X\subset \mathbb{R}^{d}$ are given.
By the above result of Park and Sandberg, there exist $k\in \mathbb{N}$, $%
w_{i}\in \mathbb{R},$ $\sigma _{i}>0$, $\mathbf{c}_{i}\in \mathbb{R}^{d}$
such that

\begin{equation*}
\left\vert f(\mathbf{x})-\sum_{i=1}^{k}w_{i}g\left( \frac{\left\Vert \mathbf{%
x}-\mathbf{c}_{i}\right\Vert }{\sigma _{i}}\right) \right\vert <\frac{%
\varepsilon }{2}\eqno(2.3)
\end{equation*}%
for all $\mathbf{x}\in X$. Note that we can write inequality (2.3) in the
form

\begin{equation*}
\left\vert f(\mathbf{x})-\sum_{i=1}^{k}g_{i}\left( \left\Vert \mathbf{x}-%
\mathbf{c}_{i}\right\Vert \right) \right\vert <\frac{\varepsilon }{2},\eqno%
(2.4)
\end{equation*}%
where $g_{i}(t):=w_{i}g(t/\sigma _{i}),$ $t\in \mathbb{R}$. Since $X$ is
compact and the distance function $\left\Vert \cdot \right\Vert $ is
continuous, the sets $\left\{ \left\Vert \mathbf{x}-\mathbf{c}%
_{i}\right\Vert :\mathbf{x}\in X\right\} $ are compact subsets of $\mathbb{R}
$; hence $\left\{ \left\Vert \mathbf{x}-\mathbf{c}_{i}\right\Vert :\mathbf{x}%
\in X\right\} \subset \lbrack a_{i},b_{i}]$ for some finite $a_{i}$ and $%
b_{i}$, $i=1,...,k$.

In 1947, Schwartz \cite{Sch} proved that continuous and $p$-th degree ($%
1\leq p<\infty $) Lebesgue integrable univariate functions are not
mean-periodic (see also \cite[Proposition 3.12]{Pin}). Note that a function $%
u\in C(\mathbb{R}^{d})$ is called mean periodic if the set $span$\ $\{u(%
\mathbf{x}-\mathbf{a}):\ \mathbf{a}\in \mathbb{R}^{d}\}$ is not dense in $C(%
\mathbb{R}^{d})$ in the topology of uniform convergence on compacta (see
\cite{Sch}). Since $g\in C(\mathbb{R})\cap L^{p}(\mathbb{R})$, by this
result of Schwartz, the set

\begin{equation*}
span\text{\ }\{g(t-\lambda ):\ \lambda \in \mathbb{R}\}
\end{equation*}%
is dense in $C(\mathbb{R)}$ in the topology of uniform convergence. This
density result means that for the given $\varepsilon $ there exist numbers $%
\rho _{ij},\lambda _{ij}\in \mathbb{R}$, $i=1,2,...,k$, $j=1,...,s_{i}$ such
that%
\begin{equation*}
\left\vert g_{i}(t)-\sum_{j=1}^{s_{i}}\rho _{ij}g(t-\lambda
_{ij})\right\vert \,<\frac{\varepsilon }{2k}\eqno(2.5)
\end{equation*}%
for all $t\in \lbrack a_{i},b_{i}]_{i},\ i=1,2,...,k.$ From (2.4) and (2.5)
it follows that

\begin{equation*}
\left\vert f(\mathbf{x})-\sum_{i=1}^{k}\sum_{j=1}^{s_{i}}\rho
_{ij}g(\left\Vert \mathbf{x}-\mathbf{c}_{i}\right\Vert -\lambda
_{ij})\right\vert <\varepsilon ,\eqno(2.6)
\end{equation*}%
for all $\mathbf{x}\in X.$ After writing the double sum in (2.6) as a single
sum we obtain the validity of (2.1).

\bigskip

\textbf{\ Corollary 2.1.} \textit{Assume $g$ is a continuous, monotone and
bounded function on $\mathbb{R}$ and $t^{d-1}g(t)$ is integrable on $%
[0,+\infty )$. Then for any continuous function $f:\mathbb{R}^{d}\rightarrow
\mathbb{R}$, for any $\varepsilon >0$ and for any compact subset $X\subset
\mathbb{R}^{d}$, there exist $m\in \mathbb{N}$, $w_{i},\nu _{i}\in \mathbb{R}
$, $\mathbf{c}_{i}\in \mathbb{R}^{d}$ such that}

\begin{equation*}
\left\vert f(\mathbf{x})-\sum_{i=1}^{m}w_{i}g\left( \left\Vert \mathbf{x}-%
\mathbf{c}_{i}\right\Vert -\nu _{i}\right) \right\vert <\varepsilon
\end{equation*}%
\textit{for all $\mathbf{x}\in X.$}

\bigskip

\textbf{Proof.} In \cite{Fun} Funahashi proved that if $g$ is a continuous,
monotone and bounded function on $\mathbb{R}$, then the function $%
h(t)=g(t+\alpha )-g(t-\alpha )$ belongs to $L_{1}(\mathbb{R})$ for any real $%
\alpha $. Thus the function $h(t)$ is not mean periodic being a continuous
and $L_{1}$ function. In addition, $t^{d-1}h(t)$ is integrable on $%
[0,+\infty )$. We can apply the above theorem to $h$ and then changing $%
h(\cdot )$ to $g(\cdot +\alpha )-g(\cdot -\alpha )$ obtain the desired
result.

\bigskip

In the following theorem, integrability condition is not required.

\bigskip

\textbf{Theorem 2.2.} \textit{Assume $g$ is a nonconstant continuous bounded
function on $\mathbb{R}$ which has a limit at infinity or minus infinity.
Then for any continuous function $f:\mathbb{R}^{d}\rightarrow \mathbb{R}$,
for any $\varepsilon >0$ and for any compact subset $X\subset \mathbb{R}^{d}$%
, there exist $m\in \mathbb{N}$, $w_{i},\nu _{i}\in \mathbb{R}$, $\mathbf{c}%
_{i}\in \mathbb{R}^{d}$ such that}

\begin{equation*}
\left\vert f(\mathbf{x})-\sum_{i=1}^{m}w_{i}g\left( \left\Vert \mathbf{x}-%
\mathbf{c}_{i}\right\Vert -\nu _{i}\right) \right\vert <\varepsilon
\end{equation*}%
\textit{for all $\mathbf{x}\in X.$}

\bigskip

\textbf{Proof.} The conditions on $g$ implies that the function $K_{0}(%
\mathbf{x})=g(\left\Vert \mathbf{x}\right\Vert )$ is a nonconstant,
continuous, bounded multivariate function. Hence $K_{0}(\mathbf{x})$ is not
a polynomial. Here we use Liao, Fang and Nuttle's result from \cite{Lia},
where they proved that for a function $K(\mathbf{x})$, which is continuous
almost everywhere, locally essentially bounded and nonpolynomial, the RBF
networks $\sum_{i=1}^{m}w_{i}K\left( \frac{\mathbf{x}-\mathbf{c}_{i}}{\sigma
_{i}}\right) $ are dense in $C(X)$ for any compact set $X\subset \mathbb{R}%
^{d}$. Note that our function $K_{0}(\mathbf{x})$ satisfy conditions of this
result.

In addition, it follows from one theorem of Schwartz (see \cite[p.907]{Sch}
and \cite[Proposition 3.12]{Pin}) that$\ g$ is not mean-periodic. Therefore,
the set $span\{g(t-\lambda ):\ \lambda \in \mathbb{R}\}$ is dense in $C[a,b]$
for any closed interval $[a,b].$ Now we have all the necessary facts to
repeat the above ideas from the proof of Theorem 2.1 and obtain the desired
final result.

\bigskip

\begin{center}
{\large \textbf{3. RBF\ networks with finitely many centroids}}
\end{center}

In this section we study approximation properties of RBF neural networks
that have a finite number of fixed centroids. We describe compact sets $%
X\subset \mathbb{R}^{d}$, for which such networks are dense in $C(X)$.

Assume we are given $k$ fixed centroids $\mathbf{c}_{1},...,\mathbf{c}_{k}$.
Put $S=\left\{ \mathbf{c}_{1},...,\mathbf{c}_{k}\right\} $. Consider the set
of RBF networks with only these centroids and arbitrary shifts

\begin{equation*}
\mathcal{G}(g,S)=\left\{ \sum_{i=1}^{m}w_{i}g\left( \left\Vert \mathbf{x}-%
\mathbf{c}\right\Vert -\nu _{i}\right) :\mathbf{c}\in S,\ w_{i},\nu _{i}\in
\mathbb{R},\ m\in \mathbb{N}\right\} .
\end{equation*}

In the above set we fix only the set $S=\left\{ \mathbf{c}_{1},...,\mathbf{c}%
_{k}\right\} $ and the activation function $g\in C(\mathbb{R})$. The
following questions arise naturally:

1) Are the RBF networks from $\mathcal{G}(g,S)$ dense in $C(\mathbb{R}^{d})$
in the topology of uniform convergence on compacta. That is, if for any
compact set $X,$ $\overline{\mathcal{G}(g,S)}=C(X)$?

2) If the answer to the above question is negative, for which compact sets $%
X\subset \mathbb{R}^{d}$, $\overline{\mathcal{G}(g,S)}=C(X)$?

\bigskip

Note that the answer to the 1st question is indeed negative. To see this,
introduce the following set of functions:

\begin{equation*}
\mathcal{R}(S)=\left\{ \sum_{i=1}^{k}g_{i}\left( \left\Vert \mathbf{x}-%
\mathbf{c}_{i}\right\Vert \right) :g_{i}\in C(\mathbb{R})\right\} .
\end{equation*}

In this set $S=\left\{ \mathbf{c}_{1},...,\mathbf{c}_{k}\right\} $ is fixed
and we vary continuous functions $g_{i}.$ Note that this is a linear space.
Since every summand $w_{i}g\left( \left\Vert \mathbf{x}-\mathbf{c}%
\right\Vert -\nu _{i}\right) $ in $\mathcal{G}(g,S)$ is a function of the
form $g_{i}\left( \left\Vert \mathbf{x}-\mathbf{c}_{i}\right\Vert \right) $
for some $\mathbf{c}_{i}$, we deduce that $\mathcal{G}(g,S)\subset \mathcal{R%
}(S).$ Thus, the set of RBF networks with fixed centroids is smaller than
the set of linear combinations of radial functions with that centroids.
Therefore, if $\mathcal{G}(g,S)$ was dense in $C(X)$, the set $\mathcal{R}%
(S) $ would be dense as well. But unfortunately, $\mathcal{R}(S)$ is not
dense in $C(X)$ for exceedingly many compact sets $X$. The reason for the
lack of density here is related to the following theorem, which is due to
Vitushkin and Henkin \cite{Vit}: For any $k$ fixed continuously
differentiable functions $h_{i},$ $i=1,...,k,$ defined on a cube $[a,b]^{d}$
the set of functions

\begin{equation*}
\left\{ \sum_{i=1}^{k}g_{i}\left( h_{i}(x_{1},...,x_{d}\mathbf{)}\right)
:g_{i}\in C(\mathbb{R})\right\}
\end{equation*}%
is nowhere dense in the space of all continuous functions on $[a,b]^{d}$
with the topology of uniform convergence. Therefore, $\mathcal{G}(g,S)$
cannot be dense in $C(X)$ if all compact sets $X$ are involved. For example,
since any set with interior contains a sufficiently small cube $[a,b]^{d}$,
it follows from the result of Vitushkin and Henkin that $\mathcal{G}(g,S)$
is not dense in $C(X)$ for any compact set $X$ with interior points. But
there still may be compact sets $X$ for which $\overline{\mathcal{G}(g,S)}%
=C(X)$ (Take, for example, a single point set $X=\{\mathbf{x}\}$) How can we
characterize such sets? To answer this question we introduce the following
objects called \textit{cycles}:

\bigskip

\textbf{Definition 3.1.} \textit{A set of points $l=\{\mathbf{x}_{1},\ldots ,%
\mathbf{x}_{n}\}\subset \mathbb{R}$\textit{$^{d}$} is called a cycle with
respect to the centroids $\mathbf{c}_{1},...,\mathbf{c}_{k}$ if there exists
a vector $\lambda =(\lambda _{1},\ldots ,\lambda _{n})\in \mathbb{Z}%
^{n}\setminus \{\mathbf{0}\}$ such that
\begin{equation*}
\sum_{j=1}^{n}\lambda _{j}\delta _{\left\Vert \mathbf{x}_{j}-\mathbf{c}%
_{i}\right\Vert }(t)=0,\ \ for~all~i=1,\ldots ,k.\eqno(3.1)
\end{equation*}%
}

\bigskip

In the above definition $\delta _{\left\Vert \mathbf{x}_{j}-\mathbf{c}%
_{i}\right\Vert }(t)$ is the characteristic function of the single point set
$\{\left\Vert \mathbf{x}_{j}-\mathbf{c}_{i}\right\Vert \}$. That is,

\begin{equation*}
\delta _{\left\Vert \mathbf{x}_{j}-\mathbf{c}_{i}\right\Vert }(t)=\left\{
\begin{array}{c}
1,\text{ if }t=\left\Vert \mathbf{x}_{j}-\mathbf{c}_{i}\right\Vert \\
0,\text{ if }t\neq \left\Vert \mathbf{x}_{j}-\mathbf{c}_{i}\right\Vert%
\end{array}%
\right. .
\end{equation*}

Let us look at Eq. (3.1) more closely. We will see that in fact it stands
for a system of simple linear equations. To understand this, fix the
subscript $i.$ Let the set $\{\left\Vert \mathbf{x}_{j}-\mathbf{c}%
_{i}\right\Vert ),$ $j=1,...,n\}$ have $s_{i}$ different values, which we
denote by $\gamma _{1}^{i},\gamma _{2}^{i},...,\gamma _{s_{i}}^{i}.$ Take
the first number $\gamma _{1}^{i}.$ Putting $t=\gamma _{1}^{i}$, we obtain
from (3.1) that

\begin{equation*}
\sum_{j}\lambda _{j}=0,
\end{equation*}%
where the sum is taken over all $j$ such that $\left\Vert \mathbf{x}_{j}-%
\mathbf{c}_{i}\right\Vert =\gamma _{1}^{i}.$ This is the first linear
equation in $\lambda _{1},...,\lambda _{n}.$ This equation corresponds to $%
\gamma _{1}^{i}$. Take now $\gamma _{2}^{i}$. By the same way, putting $%
t=\gamma _{2}^{i}$ in (3.1), we can form the second equation. Continuing
until $\gamma _{s_{i}}^{i}$, we obtain $s_{i}$ linear homogeneous equations
in $\lambda _{1},...,\lambda _{n}$. The coefficients of these equations are
the integers $0$ and $1$. By varying $i$, we finally obtain $%
s=\sum_{i=1}^{k}s_{i}$ such equations. Thus we see that (3.1), in its
expanded form, stands for the system of these linear equations. Thus the set
$l=\{\mathbf{x}_{1},\ldots ,\mathbf{x}_{n}\}$ is a cycle if this system has
a solution with nonzero integer components. In fact, it is not difficult to
understand that if the system (3.1) has a solution with nonzero real
components, then it has also a solution with nonzero integer components.
This means that in the above definition we can replace $\mathbb{Z}%
^{n}\setminus \{\mathbf{0}\}$ with $\mathbb{R}^{n}\setminus \{\mathbf{0}\}$.

We provide two simple examples of cycles here. The reader can give many
other examples easily. Assume two centroids $\mathbf{c}_{1}=(0,0)$ and $%
\mathbf{c}_{2}=(4,0)$ are given in the $xy$ plane. Then any two points $A$
and $B$ on the straight line $x=2$, which are also symmetric to the line $%
y=0 $, form a cycle. Indeed, the distances from $A$ and $B$ to $\mathbf{c}%
_{1}$ are equal and Eq. (3.1) in case of $i=1$ will be $\lambda _{1}+$ $%
\lambda _{2}=0.$ Since the distances from $A$ and $B$ to $\mathbf{c}_{2}$
are also equal, Eq. (3.1) yields the same equation for $i=2.$ Thus $\{A,B\}$
is a 2-point cycle and the vector $(\lambda _{1},\lambda _{2})$ can be taken
as $(-1,1).$ It is also easy to construct a 4-point cycle with respect to
these centroids. Consider four circles $\left\Vert \mathbf{x}-\mathbf{c}%
_{1}\right\Vert =2$, $\left\Vert \mathbf{x}-\mathbf{c}_{1}\right\Vert =3$, $%
\left\Vert \mathbf{x}-\mathbf{c}_{2}\right\Vert =4$, $\left\Vert \mathbf{x}-%
\mathbf{c}_{2}\right\Vert =3$ in the given order. These circles meet at 4
points $A,B,C,D$ in the 1-st quarter of the $xy$ plane. Each circle has only
two of these points and it is not difficult to verify that Eq. (3.1) turns
into the system

\begin{equation*}
\left\{
\begin{array}{c}
\lambda _{1}+\lambda _{2}=0 \\
\lambda _{3}+\lambda _{4}=0 \\
\lambda _{2}+\lambda _{3}=0 \\
\lambda _{1}+\lambda _{4}=0%
\end{array}%
\right.
\end{equation*}%
which has a solution $(-1,1-1,1)$. Hence, $\{A,B,C,D\}$ is a cycle with
respect to the centroids $\mathbf{c}_{1}$ and $\mathbf{c}_{2}$.

\bigskip

The second example above inspires consideration of general cycles with
respect to any given two centroids $\mathbf{c}_{1}$ and $\mathbf{c}_{2}$. In
this special case, we will use the term \textit{closed path} instead of
\textit{cycle}.

\bigskip

\textbf{Definition 3.2}. \textit{Assume $l=\left( \mathbf{x}{_{1},\mathbf{x}%
_{2},...,}\mathbf{x}_{n}\right) $ with $\mathbf{x}_{i}\neq \mathbf{x}_{i+1},$
is an ordered set with the property that $\left\Vert \mathbf{x}_{1}-\mathbf{c%
}_{1}\right\Vert =\left\Vert \mathbf{x}_{2}-\mathbf{c}_{1}\right\Vert ,$ $%
\left\Vert \mathbf{x}_{2}-\mathbf{c}_{2}\right\Vert =\left\Vert \mathbf{x}%
_{3}-\mathbf{c}_{2}\right\Vert,$ $\left\Vert \mathbf{x}_{3}-\mathbf{c}%
_{1}\right\Vert =\left\Vert \mathbf{x}_{4}-\mathbf{c}_{1}\right\Vert,...$ or
$\left\Vert \mathbf{x}_{1}-\mathbf{c}_{2}\right\Vert =\left\Vert \mathbf{x}%
_{2}-\mathbf{c}_{2}\right\Vert ,\left\Vert \mathbf{x}_{2}-\mathbf{c}%
_{1}\right\Vert =\left\Vert \mathbf{x}_{3}-\mathbf{c}_{1}\right\Vert,$ $%
\left\Vert \mathbf{x}_{3}-\mathbf{c}_{2}\right\Vert =\left\Vert \mathbf{x}%
_{4}-\mathbf{c}_{2}\right\Vert,...$ Then $l$ is called a path with respect
to the centroids $\mathbf{c}_{1}$ and $\mathbf{c}_{2}$. A path having an
even number of points $\left( \mathbf{x}_{1},\mathbf{x}_{2},...,\mathbf{x}%
_{2n}\right) $ is said to be closed if $\left( \mathbf{x}_{1},\mathbf{x}%
_{2},...,\mathbf{x}_{2n},\mathbf{x}_{1}\right) $ is also a path.}

\bigskip

Note that a closed path is a cycle. Indeed if $\left( \mathbf{x}_{1},\mathbf{%
x}_{2},...,\mathbf{x}_{2n}\right) $ is a closed path, then it is not
difficult to see that for a vector $\lambda =(\lambda _{1},\ldots ,\lambda
_{2n})$ with the components $\lambda _{i}=(-1)^{i},$ we have
\begin{eqnarray*}
\sum_{j=1}^{2n}\lambda _{j}\delta _{\left\Vert \mathbf{x}_{j}-\mathbf{c}%
_{1}\right\Vert } &=&0, \\
\sum_{j=1}^{2n}\lambda _{j}\delta _{\left\Vert \mathbf{x}_{j}-\mathbf{c}%
_{2}\right\Vert } &=&0.
\end{eqnarray*}%
Thus, by Definition 3.1, the set $\{\mathbf{x}_{1},\mathbf{x}_{2},...,%
\mathbf{x}_{2n}\}$ forms a cycle with respect to the centroids $\mathbf{c}%
_{1}$ and $\mathbf{c}_{2}$.

Cycles and paths may be defined not only for distance functions $d(\mathbf{x}%
)=\left\Vert \mathbf{x}-\mathbf{c}\right\Vert $ but also for other useful
functions too. There is a rich history of these objects defined for inner
products $\mathbf{a\cdot x}$, which were proved to be very efficient in the
theory of ridge functions (that is, functions of the form $g(\mathbf{a\cdot x%
})$). See, for example, the monograph by Ismailov \cite{Ism}.

Let $\mathbf{c}_{1},...,\mathbf{c}_{k}$ be fixed centroids and $X$ be a
compact subset of $\mathbb{R}^{d}$. For each$~i=1,\ldots ,k$, consider the
following set functions
\begin{equation*}
\tau _{i}:2^{X}\rightarrow X,\text{ }\tau _{i}(Z)=\{\mathbf{x}\in
Z:~|d_{i}^{-1}(d_{i}(\mathbf{x}))\bigcap Z|\geq 2\},
\end{equation*}%
where $d_{i}(\mathbf{x})=\left\Vert \mathbf{x}-\mathbf{c}_{i}\right\Vert $
and the symbol $|\ \ |$ denotes the cardinality of a considered set. Define $%
\tau (Z)$ to be $\bigcap_{i=1}^{k}\tau _{i}(Z)$ and define $\tau
^{2}(Z)=\tau (\tau (Z))$, $\tau ^{3}(Z)=\tau (\tau ^{2}(Z))$ and so on
inductively. Clearly, $\tau (Z)\supseteq \tau ^{2}(Z)\supseteq \tau
^{3}(Z)\supseteq ...$It is possible that for some $n$, $\tau
^{n}(Z)=\emptyset .$ In this case, one can see that $Z$ does not contain a
cycle. In general, if some set $Z\subset X$ forms a cycle, then $\tau
^{n}(Z)=Z.$ It should be remarked that the set functions $\tau _{i}$ first
appeared in Sternfeld \cite{Str}, where instead of the distance functions $%
d_{i}(\mathbf{x})$ general continuous functions are involved.

The following theorem is valid.

\bigskip

\textbf{Theorem 3.1.} \textit{Let $X$ be a compact subset of $\mathbb{R}^{d}$%
. If $\cap _{n=1,2,...}\tau ^{n}(X)=\emptyset $, then the set $\mathcal{R}%
(S) $ is dense in $C(X)$.}

\bigskip

Since functions of the form $g_{i}\left( \left\Vert \mathbf{x}-\mathbf{c}%
_{i}\right\Vert \right) $ generate a subalgebra of the space $C(X)$, Theorem
3.1 immediately follows from a general result of Sproston and Straus \cite%
{Sp} proved for a sum of continuous function algebras.

\bigskip

The following theorem establishes a sufficient condition and also a
necessary condition for the density of RBF neural networks $\mathcal{G}(g,S)$
in $C(X)$.

\bigskip

\textbf{Theorem 3.2. }\textit{Assume $g$ is a continuous $p$-th degree ($%
1\leq p<\infty $) integrable function, or $g$ is a nonconstant continuous,
bounded function, which has a limit at infinity (or minus infinity). Then
the following statements hold:}

\textit{(a) If $\bigcap _{n=1,2,...}\tau ^{n}(X)=\emptyset $, then the set $%
\mathcal{G}(g,S)$ is dense in $C(X)$;}

\textit{(b) If $\mathcal{G}(g,S)$ is dense in $C(X)$, then the set $X$ does
not contain cycles (with respect to $S$).}

\bigskip

\textbf{Proof.} (a) Suppose\textit{\ }$\bigcap\nolimits_{n=1,2,...}\tau
^{n}(X)=\emptyset $. By Theorem 3.1, the set $\mathcal{R}(S)$ is dense in $%
C(X)$. Hence for any function $f\in C(X)$ and any positive real $\varepsilon
$ there exist continuous functions $g_{i},$ $i=1,...,k$ such that

\begin{equation*}
\left\vert f(\mathbf{x})-\sum_{i=1}^{k}g_{i}\left( \left\Vert \mathbf{x}-%
\mathbf{c}_{i}\right\Vert \right) \right\vert <\frac{\varepsilon }{k+1}\eqno%
(3.2)
\end{equation*}%
for all $\mathbf{x}\in X$. Since $X$ is compact, the sets $%
Y_{i}=\{\left\Vert \mathbf{x}-\mathbf{c}_{i}\right\Vert :\ \mathbf{x}\in
X\},\ i=1,2,...,k$ are compacts as well. We know from Section 2 that the
function $g$ is not mean-periodic and hence the set

\begin{equation*}
span\text{\ }\{g(t-\theta ):\ \theta \in \mathbb{R}\}
\end{equation*}%
is dense in $C(\mathbb{R)}$ in the topology of uniform convergence on
compacta (see the proofs of Theorems 2.1 and 2.2). It follows that for the
above $\varepsilon $ there exist $c_{ij},\theta _{ij}\in \mathbb{R}$, $%
i=1,2,...,k$, $j=1,...,m_{i}$ such that%
\begin{equation*}
\left\vert g_{i}(t)-\sum_{j=1}^{m_{i}}c_{ij}g(t-\theta _{ij})\right\vert \,<%
\frac{\varepsilon }{k+1}\eqno(3.3)
\end{equation*}%
for all $t\in Y_{i},\ i=1,2,...,k.$ From (3.2) and (3.3) we obtain that

\begin{equation*}
\left\vert f(\mathbf{x})-\sum_{i=1}^{k}\sum_{j=1}^{m_{i}}c_{ij}g(\left\Vert
\mathbf{x}-\mathbf{c}_{i}\right\Vert -\theta _{ij})\right\vert <\varepsilon .%
\eqno(3.4)
\end{equation*}%
for all $\mathbf{x}\in X$. Thus $\overline{\mathcal{G}(g,S)}=C(X).$

\bigskip

(b) Suppose $\mathcal{G}(g,S)$ is dense in $C(X).$ Then for an arbitrary
positive real number $\varepsilon $, inequality (3.4) holds with some
coefficients $c_{ij},\theta _{ij}$, $i=1,2,...,k$,$\ j=1,...,m_{i}.$ Since
for each $i=1,2,...,k$, the function $\sum_{j=1}^{m_{i}}c_{ij}g(\left\Vert
\mathbf{x}-\mathbf{c}_{i}\right\Vert -\theta _{ij})$ is a function of the
form $g_{i}(\left\Vert \mathbf{x}-\mathbf{c}_{i}\right\Vert ),$ it follows
from (3.4) that the subspace $\mathcal{R}\left( S\right) $ is dense in $C(X)$%
. Let us prove that $X$ does not contain a cycle. Assume the contrary.
Assume $X$ contains a cycle, which we denote by $l=(\mathbf{x}_{1},\ldots ,%
\mathbf{x}_{n})$. Let $\lambda =(\lambda _{1},\ldots ,\lambda _{n})$ be the
vector known from Definition 3.1. Introduce the functional

\begin{equation*}
F:C(X)\rightarrow \mathbb{R},\ F(f)=\sum_{j=1}^{n}\lambda _{j}f(\mathbf{x}%
_{j}).
\end{equation*}

Clearly, $F$ is a linear bounded functional with the norm $%
\sum_{j=1}^{n}|\lambda _{j}|$. It is an exercise to check that $F(h)=0$ for
any $h\in \mathcal{R}\left( S\right) .$ By Urysohn's lemma, there exists a
continuous function $f_{0}:$ $X\rightarrow \mathbb{R}$ such that $f_{0}(%
\mathbf{x}_{j})=1$ if $\lambda _{j}>0$, $f_{0}(\mathbf{x}_{j})=-1$ if $%
\lambda _{j}<0$ and $-1<f_{0}(\mathbf{x})<1$, for any $\mathbf{x}\in
X\backslash l$. For this function, $F(f_{0})\neq 0$. We have constructed a
nonzero annihilating functional $F$. The existence of such a functional
means that $\mathcal{R}(S)$ cannot be dense in $C(X)$. The obtained
contradiction proves the 2nd statement of the theorem.

\bigskip

At the end we want to point out that the solution to the density problem for
RBF neural networks with only two fixed centroids are geometrically
explicit. In this special case, we can completely characterize all compact
sets $X\subset \mathbb{R}^{d}$ for which $\mathcal{G}(g,S)$ is dense in $%
C(X).$ To formulate the theorem, consider the following relation between
points in $X$. The relation $\mathbf{x}\thicksim \mathbf{y}$ when $\mathbf{x}
$ and $\mathbf{y}$ belong to some path in $X$ defines an equivalence
relation. The equivalence classes are called orbits (see \cite{Mar}).

The following theorem holds.

\bigskip

\textbf{Theorem 3.3.} \textit{Assume $g$ is a continuous $p$-th degree ($%
1\leq p<\infty $) integrable function, or $g$ is a nonconstant continuous,
bounded function, which has a limit at infinity (or minus infinity). Assume $X$ is a compact subset of $\mathbb{R}^{d}$
with all its orbits closed and $S=\{\mathbf{c}_{1},\mathbf{c}_{2}\}$ is
the set of fixed centroids. Then the set $\mathcal{\ G}(g,S)$ is dense in $%
C(X)$ if and only if $X$ contains no closed paths.}

\bigskip

The proof can be carried out in a similar way to the one given for the previous theorem.
Only instead of Theorem 3.1 we use the following result,
which is a corollary of the general result of Marshall and O'Farrell \cite%
{Mar} on the uniform approximation by a sum of two function algebras: If all
orbits of $X$ are closed, then for the density of $\mathcal{R}(S)$ in $C(X)$ it is
necessary and sufficient that $X$ contain no closed paths.

\bigskip

\textbf{Remark.} By definition, a closed path is a trace of some point
jumping from one position to another, alternatively on the spheres $%
\left\Vert \mathbf{x}-\mathbf{c}_{1}\right\Vert =r_{1}$, $\left\Vert \mathbf{%
x}-\mathbf{c}_{2}\right\Vert =r_{2}$ ($r_{1}$ and $r_{2}$ are not fixed),
and at the end returning to its primary position. In $\mathbb{R}^{2}$ the
circles $\left\Vert \mathbf{x}-\mathbf{c}_{1}\right\Vert =r_{1}$, $%
\left\Vert \mathbf{x}-\mathbf{c}_{2}\right\Vert =r_{2}$ form a circular
grid. Thus for density $G(g,S)$ in $C(X)$ it is necessary and sufficient
that $X$ does not contain any sequence of vertices (intersection points) $\{%
\mathbf{x}_{1},\mathbf{x}_{2},...,\mathbf{x}_{n},\mathbf{x}_{1}\}$ of this
grid with the premise that the pairs $\mathbf{x}_{i},\mathbf{x}_{i+1}$ and $%
\mathbf{x}_{i+1},\mathbf{x}_{i+2}$ lie on different circles.

\end{document}